\documentclass[runningheads]{llncs}
\usepackage{abbrv}
\usepackage[T1]{fontenc}
\usepackage{graphicx}
\usepackage{booktabs}
\usepackage{amsmath}
\usepackage{amssymb}
\usepackage{algorithm}
\usepackage{algorithmic}
\usepackage{multirow}
\usepackage[colorlinks,
            linkcolor=blue,       
            anchorcolor=blue,  
            citecolor=blue,        
            ]{hyperref}
\usepackage{cleveref}
\usepackage{bbding}
\usepackage{transparent}

\crefformat{figure}{Fig.~#2#1#3}
\crefname{figure}{Fig.}{Figs.}

\crefformat{table}{Table~#2#1#3}
\crefname{table}{Table}{Tables}

\crefformat{algorithm}{Algorithm~#2#1#3}
\crefname{algorithm}{Algorithm}{Algorithms}

\crefformat{section}{Sec.~#2#1#3}
\crefname{section}{Sec,}{Secs.}

\begin{document}
\title{Boosting Open-Vocabulary Object Detection by Handling Background Samples}
\titlerunning{Boosting OVOD by Handling Background Samples}
%
\author{Ruizhe Zeng\inst{1,2} \and
Lu Zhang\inst{1,2} \and
Xu Yang\inst{1,2} \and
Zhiyong Liu\inst{1,2,3(}\Envelope\inst{)}}

\authorrunning{R. Zeng et al.}
%
\institute{State Key Laboratory of Multimodal Artificial Intelligence Systems,\\
Institute of Automation, Chinese Academy of Sciences,\\
Beijing 100190, China\\
\email{\{zengruizhe2022,  lu.zhang, xu.yang, zhiyong.liu\}@ia.ac.cn}\and
School of Artificial Intelligence, University of Chinese Academy of Sciences,\\
Beijing 100049, China \and
Nanjing Artificial Intelligence Research of IA, Nanjing, 211100, China}
\maketitle              
%
\vspace{-10pt}
\begin{abstract}
Open-vocabulary object detection is the task of accurately detecting objects from a candidate vocabulary list that includes both base and novel categories. Currently, numerous open-vocabulary detectors have achieved success by leveraging the impressive zero-shot capabilities of CLIP. However, we observe that CLIP models struggle to effectively handle background images (\ie images without corresponding labels)  due to their language-image learning methodology. This limitation results in suboptimal performance for open-vocabulary detectors that rely on CLIP when processing background samples. In this paper, we propose Background Information Representation for open-vocabulary Detector (BIRDet), a novel approach to address the limitations of CLIP in handling background samples. Specifically, we design  Background Information Modeling (BIM) to replace the single, fixed background embedding in mainstream open-vocabulary detectors with dynamic scene information, and prompt it into image-related background representations. This method effectively enhances the ability to classify oversized regions as background. Besides, we introduce Partial Object Suppression (POS), an algorithm that utilizes the ratio of overlap area to address the issue of misclassifying partial regions as foreground. Experiments on OV-COCO and OV-LVIS benchmarks demonstrate that our proposed model is capable of achieving performance enhancements across various open-vocabulary detectors.
\end{abstract}

\section{Introduction}
\label{sec:intro}
Open-Vocabulary Object Detection (OVOD) is a challenging task that aims to break the closed-world assumption of object detection, \ie, detectors should be capable of detecting not only in base categories but also novel categories~\cite{zareian2021open}. Compared to closed-world detectors~\cite{ren2015faster,redmon2016you}, open-vocabulary detectors are more suitable for real-world applications like robotics~\cite{geiger2013vision} and autonomous driving~\cite{dollar2009pedestrian}, as novel objects frequently appear in these scenarios.

\begin{figure}
    \centering
    \includegraphics[width=0.8\textwidth]{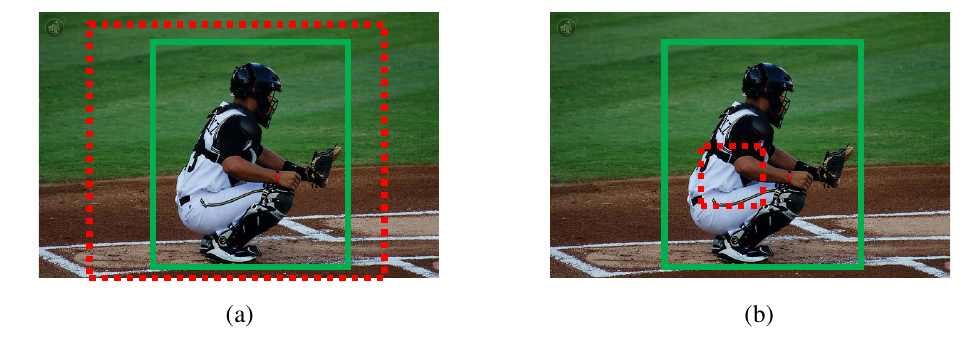}
      \caption{(a) An example of \textbf{oversized regions} and (b) an example of \textbf{partial regions}. Green boxes are ground truths, while red dashed boxes are oversized regions and partial regions respectively. Oversized regions have excessive background information, while partial regions only contain part of objects.}
    \label{fig:1}
\end{figure}

Recently, pre-trained vision-language models (\eg CLIP~\cite{radford2021learning}) perform well in zero-shot image classification by aligning image-text features, and they are adopted by numerous open-vocabulary detectors~\cite{zareian2021open,zhong2022regionclip,gu2021open}. However, CLIP employs contrastive learning as its language-image learning method, which implies that each training image has a corresponding label and background images are excluded during training.  Consequently, the ability of CLIP to process background images warrants further investigation, especially when applied in the OVOD task, as it typically involves a substantial number of background region samples. Typically, there are two types of background region samples in the OVOD task, as shown in \cref{fig:1}: (1) regions containing excessive background information, which we refer them as \textbf{oversized regions}, and (2) regions containing only partial objects, which we refer them as \textbf{partial regions}. During inference, when the intersection-over-unions (IoUs) of these samples with ground truths fall below a certain threshold, they should be classified as background.

\begin{figure}
    \centering
    \includegraphics[width=\textwidth]{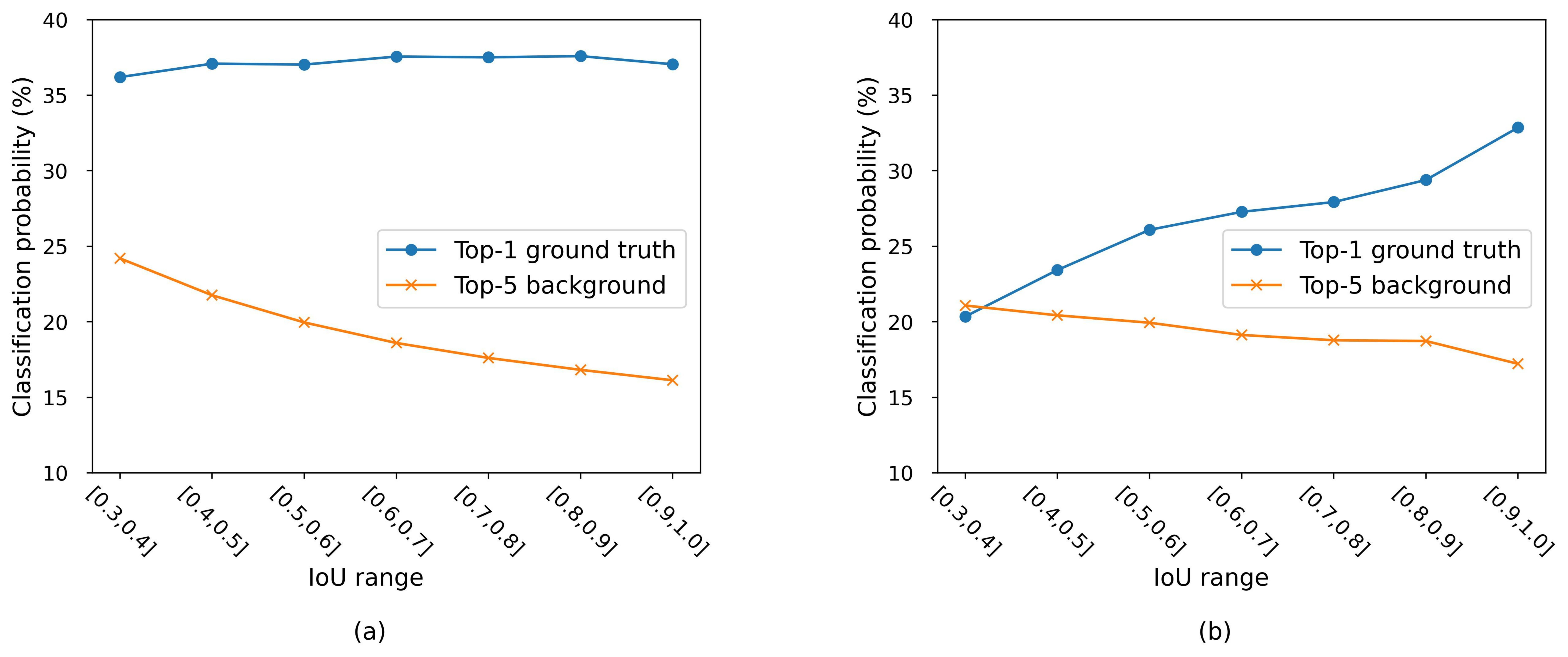}
      \caption{(a) Classification results on oversized regions of the CLIP RN50$\times$64 model~\cite{radford2021learning}. (b) Classification results on partial regions of the CLIP RN50$\times$64 model. The x-axis stands for the IoU range of regions, while the y-axis stands for classification probabilities. We measure the top-1 classification probability for ground truth categories and top-5 background classification probabilities respectively. }
    \label{fig:2}
\end{figure}

To assess CLIP's ability on classifying these background region samples, we conducted a preliminary experiment on the MS COCO validation dataset~\cite{lin2014microsoft}. We sampled oversized regions and partial regions according to different IoU ranges on the dataset and fed them into CLIP for classification. Like~\cite{zareian2021open,kuo2022open}, we incorporated the ``background'' word embedding and zero embedding to represent background information. The probabilities for top-1 classification probability for ground truth categories and top-5 classification probability of the background information were respectively calculated. As illustrated in \cref{fig:2},  the classification results on oversized regions are relatively similar across different IoUs, indicating that CLIP has limited capability in modeling background information.  As for partial regions, we observe that CLIP often erroneously classifies them as foreground objects even when they have relatively low IoU with ground truths. Since novel objects are inaccessible during training, most of current open-vocabulary detectors directly migrate the zero-shot capabilities of CLIP to handle novel objects~\cite{gu2021open,lin2022learning}. However, they often neglect the aforementioned limitations of CLIP, which may result in a substantial number of false positive (FP) detection results.

To address the aforementioned issues, in this paper, we propose \textbf{B}ackground \textbf{I}nformation \textbf{R}epresentation for Open-vocabulary \textbf{Det}ector (BIRDet), a simple but effective plug-and-play approach that handles background region samples in open-vocabulary detectors. We observe that oversized regions typically contain numerous elements of the image scene, inspiring us to use semantic information of scenes to model the background. Hence, we add a \textbf{B}ackground \textbf{I}nformation \textbf{M}odeling (BIM) branch to open-vocabulary detectors. Specifically, considering the complex and diverse background information present in an image, we select the top background recognition results and use them to reduce the misclassification of oversized regions. By incorporating more generalized background information compared to a simple, fixed embedding, BIM can also mitigate the erroneous classification of novel objects as background. Besides, we design \textbf{P}artial \textbf{O}bject \textbf{S}uppression (POS) algorithm to handle partial regions. Unlike traditional Non-Maximum Suppression (NMS)~\cite{neubeck2006efficient}, we use the ratio of overlap area to region area to determine whether regions should be suppressed. It can effectively remove partial regions without adversely affecting the detection of occluded objects.

Our contributions are summarized as follows:
\begin{itemize}
    \item We propose a background information modeling module for open-vocabulary object detection. It extracts background information from images to mitigate the misclassification of oversized regions, thereby enhancing the detection of novel categories.

    \item We propose a novel partial object suppression algorithm to effectively reduce false positive detection results caused by partial regions.

    \item Experimental evidence shows that our plug-and-play model can significantly reduce detection error on novel categories when combined with various open-vocabulary detectors, and get better results on several benchmarks including OV-COCO~\cite{lin2014microsoft} and OV-LVIS~\cite{gupta2019lvis}.
\end{itemize}

\section{Related Work}
\subsubsection{Open-vocabulary Object Detection.}  Currently, numerous open-vocabulary detectors use CLIP~\cite{radford2021learning} to achieve region-text alignment. ViLD~\cite{gu2021open} and HierKD~\cite{ma2022open} distill  CLIP features to object detectors~\cite{gou2021knowledge},  enabling them to leverage CLIP's zero-shot capability.  RegionCLIP~\cite{zhong2022regionclip} and CLIPSelf~\cite{wu2023clipself} focus on differences in visual information between classification and detection tasks and fine-tune CLIP for better region-text alignment. Detic~\cite{zhou2022detecting} employs pseudo-labels generation methods to augment region-text datasets, thus expanding available region-text information during training. However, most of them neglect to address the weaknesses in processing background region samples by CLIP, particularly oversized regions and partial regions in object detection tasks. Though LP-OVOD~\cite{pham2024lp} identifies the limitations of CLIP and employs a sigmoid classifier, its performance is contingent upon the generated pseudo-labels, which are often unreliable.

\subsubsection{CLIP for Out-of-distribution (OOD) Detection.}  OOD detection aims to identify images that belong to categories not present in the training dataset and usually originate from a different distribution~\cite{yang2021generalized}. Numerous OOD models~\cite{shu2023clipood,wang2023clipn,esmaeilpour2022zero} are conducted on CLIP~\cite{radford2021learning} as it demonstrates good generalization capability on visual-language information. CLIPN~\cite{wang2023clipn}  proposes a ``no'' prompt strategy to teach CLIP how to distinguish between correct and incorrect semantic matching with images. To fully unleash the potential of CLIP’s ability,  LoCoOp~\cite{miyai2024locoop} inherits the optimization approach of CoOp~\cite{zhou2022learning} to CLIP, using CLIP's local features as OOD features to achieve the optimization of CLIP on OOD tasks. These methods are applied to image classification tasks, while our method handles background samples in object detection tasks.

\section{Method}

\subsection{Preliminaries}

\begin{figure}[h]
  \centering
  \includegraphics[width=\textwidth]{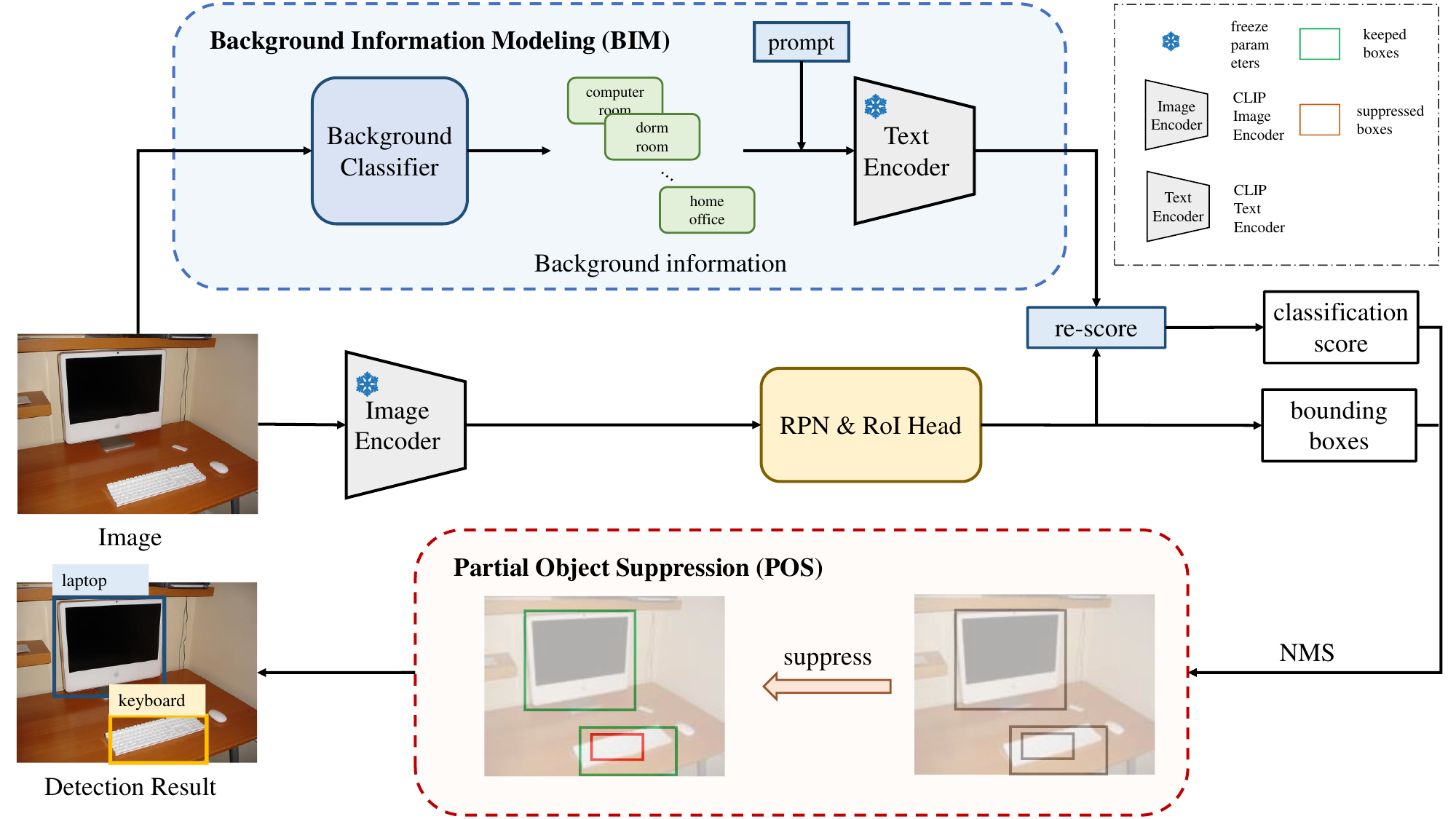}
  \caption{Method overview. It consists of two novel modules: Background Information Modeling (BIM) and Partial Object Suppression (POS) algorithm. In addition to extracting foreground features using the  CLIP image encoder, it classifies images into scene information and prompt it as background representations. After encoding, we use it as the background embedding for the RoI head and re-score  classification scores to reduce bias caused by scene information. When initial detection results are available, we use the POS algorithm to suppress partial regions and obtain the  final detection results.}
  \label{fig:3}
\end{figure}

\subsubsection{CLIP.} CLIP is a large-scale visual-language model based on contrastive learning~\cite{radford2021learning}. Specifically, CLIP  compiles a dataset containing 400 million images from the internet, with each image accompanied by a corresponding text description. During training, given a batch of $N$ image-text pairs, CLIP maximizes the similarity between  $N$ paired  image-text embeddings, while minimizing the similarity between $N^2 - N$ mismatched image-text embeddings,  The loss function is formulated as follows to achieve feature alignment between images and texts:

\begin{equation}
\mathcal{L} = \frac{1}{2N} \sum_{i=1}^N \left[ -\log \frac{\exp(s_{ii} / \tau)}{\sum_{j=1}^N \exp(s_{ij} / \tau)} - \log \frac{\exp(s_{ii} / \tau)}{\sum_{j=1}^N \exp(s_{ji} / \tau)} \right], 
\end{equation}
where $s_{ij}$ is the similarity between $i$-th image and $j$-th text, $\tau$ is the temperature parameter. 

However, we observe that during training, all images have corresponding caption labels and background images are not included. Consequently, CLIP lacks the capability to effectively classify background samples encountered during detection, as these samples do not have corresponding text labels.

\subsubsection{Problem Setup.} Following the settings of~\cite{gu2021open,zhou2022detecting}, open-vocabulary detectors can leverage instance-level supervised datasets with base category $C_B$ and image-level supervised datasets at the training stage. During inference, detectors should detect objects belonging to base category $C_B$ or novel category $C_N$, where $C_B \bigcap C_N = \emptyset$. Generally, detectors should classify background region samples as $C_{\text{bg}}$.

Most of open-vocabulary detectors follow the two-stage detection frameworks, such as the popular Mask-RCNN~\cite{he2017mask}. Given a RGB image $I \in \mathbb{R}^{H \times W \times 3}$, two-stage detectors typically use region proposal network (RPN~\cite{ren2015faster}) to generate region proposals for $I$ after extracting features, and use region-of-interest (RoI) head to predict location and classification results. In order to reduce the impact on accuracy caused by false-positive (FP) results, these detectors typically employ the NMS algorithm~\cite{neubeck2006efficient} to remove candidate FP results.

\subsubsection{Method Overview.}~\Cref{fig:3} presents an overview of our BIRDet model. We adopt a two-stage object detector framework~\cite{he2017mask,ren2015faster} and integrate CLIP model to build the baseline detector. Apart from extracting foreground features, we also extract background features from images, classify them into scene information, and prompt it as background representations (\cref{sec:Background modeling module}). Since scene information is more similar to object categories than fixed embeddings, we re-score classification results from the RoI head. When initial detection results are available, we apply the partial object suppression algorithm to address CLIP's limitations in handling partial regions (\cref{sec:Small Object Suppression}). The aforementioned methods can effectively reduce false positive detection results and enhance the detection of novel categories. We will elaborate on these methods in detail in the subsequent sections.

\subsection{Background Information Modeling}
\label{sec:Background modeling module}

In this section, we introduce \textbf{B}ackground \textbf{I}nformation \textbf{M}odeling (BIM), a framework to construct background feature representations for open-vocabulary detectors. Unlike most open-vocabulary detectors that use a single, fixed embedding for background representation~\cite{zhong2022regionclip,du2022learning}, BIM constructs background embedding for each image based on its scene information, thereby alleviating the weakness of CLIP in identifying oversize regions as background.

\subsubsection{Training BIM.} As BIM has different feature concerns from object detection and image captioning~\cite{xie2020scene}, which needs to focus on background information rather than foreground objects.
We use an additional image-level supervised dataset $D_s$ to train BIM, which is more focused on background-level information and has low label cost.  Specifically, we extract the set of scene information $C_s = \{d_1,d_2,\cdots,d_S\}$ ($S$ is the number of scene information) from $D_s$  and employ a small CNN-based network~\cite{he2016deep} to match images with corresponding scene information. We use cross-entropy loss to train the classifier in BIM and keep other model weights frozen. Note that this process is similar to image classification tasks, so using models pre-trained on $D_s$ is also appropriate.

\subsubsection{Using BIM for Open-vocabulary Detector.}  Given an image $I$ and BIM module $\mathcal{F}(\cdot)$, we can extract probability distribution $P_s = \mathcal{F}(I) \in R^S$ on $D_s$. As real-world images may have various scene representations, We extract $K$ most probable scene information $\{d_{s_1}, d_{s_2}, \cdots, d_{s_K} \}$ for background modeling and prompt them as $\{p(d_{s_1}), p(d_{s_2}), \cdots, p(d_{s_K}) \}$ before sending them to CLIP text encoder. Specifically, we calculate background  embedding $t_{bg}$ of $I$ as follows:

\begin{equation}
t_{bg}(I) = \frac{1}{K} \sum_{k=1}^K \mathcal{T} (p(d_{s_k})),
\end{equation}
where $\mathcal{T}(\cdot)$ means CLIP text encoder. We concatenate class embeddings ($C_B$ when training and $C_B \bigcup C_N$ when inference) and $t_{bg}(I)$ to classify region proposals.

However, as scene information has co-occurrence relationships with foreground objects~\cite{liu2018structure},  $t_{bg}(I)$ usually has higher cosine similarities with object categories in $I$ than traditional background embeddings used by~\cite{zhong2022regionclip,du2022learning}. For example, the category name ``sink'' is closer to scene information ``bathroom'' than all-zero embedding in feature space, and this phenomenon may introduce classification bias during inference. To mitigate it,  we re-score classification results from RoI head by using scene-object similarities. Specifically, for the $i$-th category in $C_B \bigcup C_N$, we calculate the cosine similarity $\phi_{i,bg}$ between its embedding and $t_{bg}(I)$. Inspired by~\cite{fang2024simple}, we normalize the similarity score as $\widetilde{\phi}_{i,bg}$ and use the sigmoid function to calculate re-score coefficient $r_{i}$ as follows:

\begin{equation}
r_{i} = sigmoid(\widetilde{\phi}_{i,bg}).
\end{equation}
Finally, when the classification score of the $j$-th initial result is available, we get the final classification result for the $i$-th category as follows:

\begin{equation}
s_{i,j} = (s^{\text{initial}}_{i,j})^{1-\alpha}r_i^{\alpha},
\end{equation}
where $\alpha$ is a hyperparameter, $s^{\text{initial}}_{i,j}$ is the initial classification score and $s_{i,j}$ is the final classification score.

\subsection{Partial Object Suppression}
\label{sec:Small Object Suppression}

In this section, we introduce  \textbf{P}artial \textbf{O}bject \textbf{S}uppression (POS), a training-free algorithm designed to remove partial regions. Most open-vocabulary detectors use the non-maximum suppression (NMS) algorithm to postprocess detection results~\cite{neubeck2006efficient}. NMS employs IoU-based suppression method, which cannot effectively remove background examples of partial regions as these regions may have low IoU with other bounding boxes. A novel approach is directly removing small bounding boxes with low confidence. However, the confidence scores for $C_N$ are generally low, as instances of $C_N$ are not involved during training. Moreover, such an approach would affect the detection of occluded objects. Hence, we use the \textbf{O}verlap \textbf{A}rea \textbf{R}atio (OAR) as the criterion to suppress partial regions. It is worth noting that we mainly use POS on $C_N$ as open-vocabulary detectors can already handle partial regions of $C_B$ through the training process.

Specifically, given two bounding boxes $b_1, b_2$ of the same category, we compute OAR as follows:

\vspace{-10pt}
\begin{equation}
\text{OAR}(b_1, b_2) = \frac{|b_1 \bigcap  b_2|}{|b_1|}.
\end{equation}
We notice that $\text{OAR}(b_1, b_2)$ is not necessarily equal to $\text{OAR}(b_2, b_1)$ since we only need to consider whether to suppress the smaller bounding box. 

\begin{figure}[h]
  \centering
  \includegraphics[width=\textwidth]{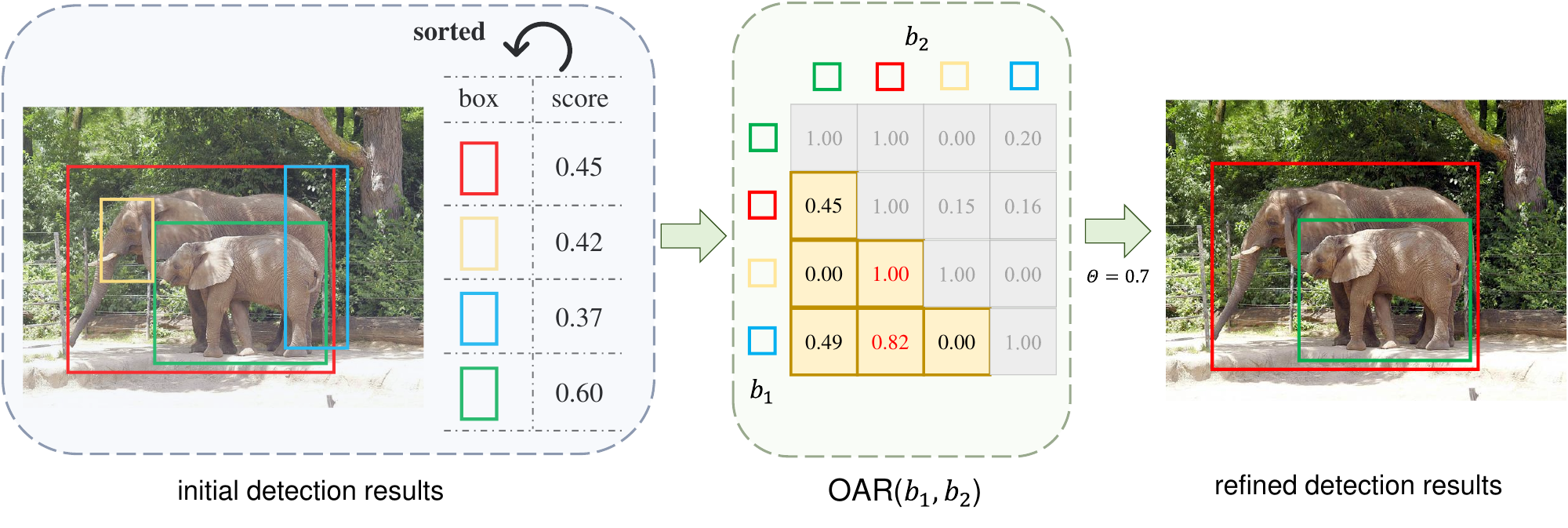}
  \caption{An example of the partial object suppression (POS) algorithm. After NMS, we sort candidate bounding boxes by scores and compute the overlap area ratio (OAR) between the boxes. We suppress boxes with  OAR no less than $\Theta$ to obtain the refined detection results.}
  \label{fig:4}
\end{figure}

Then, for every category $c_i \in C_N$, when we have corresponding initial detection results $\{b_j,s_j\}_{j=1}^{N_i}$ (where $b_j,s_j$ are the bounding box and confidence score respectively, $N_i$ is the number of initial detection results of $c_i$), we can use~\cref{alg: SOS} to refine the detection result.~\Cref{fig:4} shows an example of POS. Similar to NMS~\cite{neubeck2006efficient}, we sort candidate bounding boxes by their confidence scores, if a bounding box $b$ has  OAR with one of the selected bounding boxes $b_{\text{selected}}$ that is no less than the predefined threshold $\Theta$, then the corresponding result for $b$ will be discarded. Compared to partial regions, occluded objects typically have low OAR with other bounding boxes even though they are also small in size. Therefore, the POS algorithm will not affect the detection of occluded objects.

\begin{algorithm}
\caption{Partial Object Suppression (POS)}
\label{alg: SOS}
\begin{algorithmic}[1]
\STATE \textbf{Input:} bounding boxes $\mathcal{B} = \{b_1,\cdots,b_{N_i}\}$, corresponding confidence scores $\mathcal{S} = \{s_1,\cdots,s_{N_i}\}$, predefined threshold $\Theta$
\STATE \textbf{Output:} selected bounding boxes $\mathcal{B}_S$ and corresponding confidence scores $\mathcal{S}_S$

\STATE $\mathcal{B}_S \gets \{\}$
\WHILE{$\mathcal{B} \neq \emptyset$}
    \STATE $m \gets \arg\max \mathcal{S}$
    \STATE $\mathcal{M} \gets b_m$
    \STATE $\mathcal{B}_S \gets \mathcal{B}_S \cup \{\mathcal{M}\}$
    \STATE $\mathcal{B} \gets \mathcal{B} - \{\mathcal{M}\}$; $\mathcal{S} \gets \mathcal{S} - \{s_m\}$
    \FOR{$b_j$ \textbf{in} $\mathcal{B}$}
        \IF{\text{OAR}($b_j$, $\mathcal{M}) \ge \Theta$}
            \STATE $\mathcal{B} \gets \mathcal{B} - \{b_j\}$; $\mathcal{S} \gets \mathcal{S} - \{s_j\}$
        \ENDIF
    \ENDFOR
\ENDWHILE
\end{algorithmic}
\end{algorithm}

\section{Experiments}

\subsection{Experiment Setup}

\subsubsection{Datasets and Evaluation Metrics.} Following~\cite{gu2021open}, we conduct experiments on OV-COCO and OV-LVIS benchmarks respectively. OV-COCO splits MS COCO dataset~\cite{lin2014microsoft} into 48 base categories and 17 novel categories. OV-LVIS takes 866 frequent and common categories in the LVIS dataset~\cite{gupta2019lvis} for training and takes 337 rare categories as novel categories.  For OV-COCO, We adopt box AP at IoU threshold 0.5 on each category and compute the mean AP on base categories, novel categories, and all categories respectively (\ie  $\text{mAP}_{50}^{\text{base}}$, $\text{mAP}_{50}^{\text{novel}}$ and $\text{mAP}_{50}^{\text{all}}$).  For OV-LVIS, we compute box AP on frequent, common, rare, and all categories respectively (\ie  $\text{mAP}_{\text{f}}$, $\text{mAP}_{\text{c}}$ and $\text{mAP}_{\text{r}}$, $\text{mAP}_{\text{all}}$). Following \cite{gu2021open}, we mainly focus on $\text{mAP}_{50}^{\text{novel}}$ and $\text{mAP}_{\text{r}}$ to evaluate the generalization ability on novel categories.

\subsubsection{Implementation Details.} We conduct experiments on open-vocabulary detectors following the F-VLM framework~\cite{kuo2022open}, where detectors use the original CLIP model or the finetuned CLIP model~\cite{radford2021learning} as the backbone, and adjust the parameters of other modules during training. We follow all the hyperparameters of F-VLM for our baseline except training for 48 epochs and 8 batch sizes on OV-LVIS~\cite{gupta2019lvis} due to computational constraints. For BIM, we use Places365~\cite{zhou2017places} as the additional image-supervised dataset. Places365 contains 10 million images with 365 scene categories covering almost all common scenarios. We use a pre-trained ResNet18 model~\cite{he2016deep} on Places365 for BIM and use $K=5$ most probable scene information. As oversized regions usually contain only part of the image scene rather than the whole scene, we use    ``\textit{Part of \{scene information\}}'' as the background prompt. We use $\alpha=0.2$ for the re-score module to alleviate the bias caused by scene information. For the POS algorithm, we set $\Theta=0.5$ to suppress partial regions. As BIRDet is a plug-and-play model, we build BIRDet on several open-vocabulary detectors and adopt the same hyperparameter settings as them, except for the classification loss weight since the background embedding under BIM is more similar to object categories than a single, fixed embedding. The specific settings will be clarified in the subsequent explanations. We use mmdetection\cite{chen2019mmdetection} to build our model and use two NVIDIA GeForce RTX 3090 as our GPU experimental conditions.

\subsection{Main Results}

\begin{table}
    \caption{Comparison of various open-vocabulary object detectors on the OV-COCO benchmark.}
    \label{tab1:OVCOCO}
    \centering
    \begin{tabular}{l|c|ccc}
\hline
\makebox[0.12\textwidth][l]{Method}          & \makebox[0.12\textwidth][c]{Backbone}  & \makebox[0.13\textwidth][c]{$\text{mAP}_{50}^{\text{novel}}$}    & \makebox[0.13\textwidth][c]{$\text{mAP}_{50}^{\text{base}}$}    & \makebox[0.13\textwidth][c]{$\text{mAP}_{50}^{\text{all}}$}      \\
\hline
ViLD~\cite{gu2021open}              & RN50                           & 27.6  & \transparent{0.4}{59.5} & \transparent{0.4}{51.2}  \\
Detic~\cite{zhou2022detecting}             & RN50             & 27.8  & \transparent{0.4}{51.1} & \transparent{0.4}{45.0}    \\
OV-DETR~\cite{zang2022open}           & RN50                            & 29.4  & \transparent{0.4}{61.0}   & \transparent{0.4}{52.7}  \\
RegionCLIP~\cite{zhong2022regionclip}        & RN50              & 31.4  & \transparent{0.4}{57.1} & \transparent{0.4}{50.4}  \\
VLDet~\cite{lin2022learning}             & RN50             & 32.0    & \transparent{0.4}{50.6} & \transparent{0.4}{45.8}  \\

LP-OVOD~\cite{pham2024lp}           & RN50                          & 40.5  & \transparent{0.4}{60.5} & \transparent{0.4}{55.2}  \\
\hline
baseline        & RN50             & 27.8  & \transparent{0.4}{46.9} & \transparent{0.4}{41.9}  \\
BIRDet & RN50     &29.8($\uparrow$2.0)       &\transparent{0.4}{50.0}      &\transparent{0.4}{44.7}       \\
\hline
BARON~\cite{wu2023aligning}             & RN50                             & 34.0    & \transparent{0.4}{60.4} & \transparent{0.4}{53.5}  \\
BARON+BIRDet      & RN50                 &  34.6($\uparrow$0.6)    &  \transparent{0.4}{50.7}    &  \transparent{0.4}{46.7}      \\
\hline
CLIM~\cite{wu2024clim}              & ViT-B            &  29.7     &   \transparent{0.4}{50.0}   &   \transparent{0.4}{44.7}    \\
CLIM+BIRDet       & ViT-B    &    32.6($\uparrow$2.9)   &   \transparent{0.4}{47.1}   &  \transparent{0.4}{43.3}     \\
\hline
CLIPSelf~\cite{wu2023clipself}          & ViT-B           & 37.6  & \transparent{0.4}{54.9} & \transparent{0.4}{50.4}  \\
CLIPSelf+BIRDet   & ViT-B     &  40.5($\uparrow$2.9)     &  \transparent{0.4}{54.8}    &  \transparent{0.4}{51.0}     \\
CLIPSelf          & ViT-L              & 44.3  & \transparent{0.4}{64.1}
 & \transparent{0.4}{59.0}    \\
CLIPSelf+BIRDet   & ViT-L    &   \textbf{46.2}($\uparrow$1.9)    &  \transparent{0.4}{63.0}    &  \transparent{0.4}{58.6}    \\
\hline
\end{tabular}
\end{table}

\subsubsection{OV-COCO Benchmark.}~\cref{tab1:OVCOCO} shows the main results of our method on the OV-COCO benchmark. We build our baseline following F-VLM~\cite{kuo2022open} and adopt BIRDet on it. Results show that BIRDet can bring 2.0 $\text{mAP}_{50}^{\text{novel}}$ improvements. Then, we investigate two prevalent F-VLM-style open-vocabulary models: CLIM~\cite{wu2024clim} and  CLIPSelf~\cite{wu2023clipself}. We follow their experimental settings except for using 0.6, 0.2 and 0.6 background classification loss weights for CLIM,  ViT-B~\cite{dosovitskiy2020image} CLIPSelf and ViT-L~\cite{dosovitskiy2020image} CLIPSelf respectively. Experimental results demonstrate that BIRDet can improve 1.9$\sim$2.9 $\text{mAP}_{50}^{\text{novel}}$ across these methods. We also conduct experiments on BARON~\cite{wu2023aligning}, which finetunes its backbone during training. Our method also achieves  0.6 $\text{mAP}_{50}^{\text{novel}}$ improvements, demonstrating a certain level of effectiveness of BIRDet even when applied within a non-frozen backbone framework.
\vspace{-10pt}
\subsubsection{OV-LVIS Benchmark.}~\cref{tab2:OVLVIS} shows the main results of our method on the OV-LVIS benchmark. As the box APs on OV-LVIS were not reported in original papers, the results of CLIM~\cite{wu2024clim} and CLIPSelf~\cite{wu2023clipself} are reproduced results. We use 0.9 and 0.7 background classification loss weights for CLIM and ViT-B CLIPSelf respectively. Experimental results indicate that our method can make certain improvements on CLIM and CLIPSelf, however, these improvements are relatively smaller than those on OV-COCO. We believe this could be due to two reasons. The first is that, through preliminary experiments, we observe that the primary source of misclassification in OV-LVIS is detecting bounding boxes as wrong categories. Our approach mainly focuses on the relationship between category information and background information, making limited improvements in addressing category relationships. The second is that, due to the large number of categories in the LVIS dataset, the background information generated by BIM is more likely to be similar to certain categories. Therefore, bias caused by scene information persists even after re-scoring the classification results. A straightforward idea is to use the scene information and object co-occurrence relationships generated by BIM to further enhance the effect of BIRDet, and we plan to address this issue in our future work.

\begin{table}
    \caption{Comparison of various open-vocabulary object detectors on the OV-LVIS benchmark.}
    \label{tab2:OVLVIS}
    \centering
    \begin{tabular}{l|c|cccc}
\hline
\makebox[0.11\textwidth][l]{Method}          & \makebox[0.12\textwidth][c]{Backbone}  & \makebox[0.1\textwidth][c]{$\text{mAP}_{\text{r}}$}    & \makebox[0.1\textwidth][c]{$\text{mAP}_{\text{c}}$}    & \makebox[0.1\textwidth][c]{$\text{mAP}_{\text{f}}$}    & \makebox[0.1\textwidth][c]{$\text{mAP}_{\text{all}}$}  \\
\hline
ViLD~\cite{gu2021open}            & RN50                     & 16.7 & \transparent{0.4}{26.5} & \transparent{0.4}{34.2} & \transparent{0.4}{27.8} \\

RegionCLIP~\cite{zhong2022regionclip}      & RN50               & 17.1 & \transparent{0.4}{27.4} & \transparent{0.4}{34.0} & \transparent{0.4}{28.2} \\

BARON~\cite{wu2023aligning}           & RN50                   & 20.1 & \transparent{0.4}{28.4} & \transparent{0.4}{32.2} & \transparent{0.4}{28.4} \\

DetPro~\cite{du2022learning}          & RN50                     & 20.8 & \transparent{0.4}{27.8} & \transparent{0.4}{32.4} & \transparent{0.4}{28.4} \\
LBP~\cite{Li_2024_CVPR}           & RN50             & 22.2 & \transparent{0.4}{28.8} & \transparent{0.4}{32.4} & \transparent{0.4}{29.1} \\
CORA~\cite{wu2023cora}           & RN50$\times$4             &22.2 & \transparent{0.4}{-} & \transparent{0.4}{-} & \transparent{0.4}{-} \\
\hline
baseline        & RN50                   &  18.0    & \transparent{0.4}{24.0}     &  \transparent{0.4}{28.3}    &  \transparent{0.4}{24.5}    \\
BIRDet          & RN50              &  18.8($\uparrow$0.8)    &  \transparent{0.4}{23.3}    & \transparent{0.4}{27.8}     & \transparent{0.4}{24.3}     \\
\hline
CLIM~\cite{wu2024clim}            & ViT-B               & 25.1     & \transparent{0.4}{32.3}     &\transparent{0.4}{34.4}      &\transparent{0.4}{31.9}      \\
CLIM+BIRDet     & ViT-B                 & 25.4($\uparrow$0.3)     & \transparent{0.4}{31.7}     &\transparent{0.4}{33.6}      &\transparent{0.4}{30.6}      \\
\hline
CLIPSelf~\cite{wu2023clipself}        & ViT-B                     & 25.8 & \transparent{0.4}{22.1} & \transparent{0.4}{31.5} & \transparent{0.4}{26.4} \\
CLIPSelf+BIRDet & ViT-B                & \textbf{26.0}($\uparrow$0.2)     & \transparent{0.4}{21.7}    & \transparent{0.4}{29.5}     & \transparent{0.4}{25.5}     \\
\hline
\end{tabular}

\end{table}

\subsection{Ablation Study}

In this section, we conduct ablation studies on OV-COCO benchmarks. We use ViT-B CLIPSelf~\cite{wu2023clipself} as the  base open-vocabulary detector and evaluate our method on it.

\subsubsection{Effectiveness of Our Proposed Modules.} We examine the effectiveness of our different modules and algorithms. \Cref{tab3:allab} shows our promising results on top of ViT-B CLIPSelf~\cite{wu2023clipself} detector. Results show that BIM can utilize scene information and re-score classification results to improve detection accuracy for novel categories. However, the BIM module is effective primarily for oversized regions and has a relative weak effect on partial regions as they usually contain little background information. For POS algorithm, it significantly reduces false positives caused by partial regions during inference and enhances mAP for novel categories, despite its simplicity. Nevertheless, the POS algorithm struggles to suppress oversized regions, which often have a low OAR with true positive samples (typically equal to IoU). In summary, our designed modules exhibit both strengths and weaknesses, and they can complement each other effectively when integrated.

\begin{table}
\centering
\caption{Ablation study on our proposed modules in BIRDet. SI means using scene information as background embedding.}
\label{tab3:allab}
\begin{tabular}{ccc|c}
\hline
\multicolumn{2}{c}{BIM} & \makebox[0.11\textwidth]{\multirow{2}{*}{POS}} & \makebox[0.11\textwidth]{\multirow{2}{*}{$\text{mAP}_{50}^{\text{novel}}$}}  \\
\makebox[0.1\textwidth][c]{SI}      & \makebox[0.1\textwidth][c]{Re-score}      &                      &                          \\
\hline
        &               &                      & 37.5                                \\
$\surd$       &               &                      & 38.8                                   \\
$\surd$       & $\surd$              &                      &    39.0                                   \\
        &               & $\surd$                     & 39.9                                  \\
$\surd$        &               & $\surd$                     & 40.2                                  \\
$\surd$        & $\surd$              & $\surd$                     & \textbf{40.5}                  \\
\hline
\end{tabular}
\end{table}

\subsubsection{Study of Top-$K$ Scene Information.} We examine the optimal number for the selection of top-$K$ scene information. As illustrated in \cref{tab5:topk}, low $K$ values (\eg $K=1$) may result in a partial representation of scene information understanding from BIM, while high $K$ values (\eg $K=10$) may lead to inaccurate outputs of scene information. We choose $K=5$ for the BIM module to balance these two unfavorable situations.

\subsubsection{Study on the Exponent Coefficient $\alpha$ for Re-score.} We investigate the geometric mean exponent coefficient $\alpha$ that best suits the BIM module. As shown in \cref{tab4:rescore}, we select $\alpha=0.2$ to minimize the bias brought by scene information, while also avoiding significant interference with the detection results of the open vocabulary detector itself.

\begin{minipage}{\textwidth}
\begin{minipage}[t]{0.48\textwidth}
\makeatletter\def\@captype{table}
\caption{Effectiveness of  top-$K$ \\scene information selection.\\}
\label{tab5:topk}
    \begin{tabular}{c|ccc}
    \hline
    \makebox[0.16\textwidth][c]{$K$}     
    & \makebox[0.16\textwidth][c]{1}       
    & \makebox[0.16\textwidth][c]{5}      
    & \makebox[0.16\textwidth][c]{10}     \\ 
    \hline
    $\text{mAP}_{50}^{\text{novel}}$  
    &36.3
    &\textbf{40.5}
    &39.0\\
    \hline
\end{tabular}
\end{minipage}
\begin{minipage}[t]{0.48\textwidth}
\makeatletter\def\@captype{table}
\centering
\caption{Effectiveness of coefficient $\alpha$ for re-score.\\}
\label{tab4:rescore}
\begin{tabular}{c|ccccc}
    \hline
    \makebox[0.14\textwidth][c]{$\alpha$} & \makebox[0.1\textwidth][c]{0.0} & \makebox[0.1\textwidth][c]{0.1} & \makebox[0.1\textwidth][c]{0.2} & \makebox[0.1\textwidth][c]{0.3} & \makebox[0.1\textwidth][c]{0.4} \\
    \hline
     $\text{mAP}_{50}^{\text{novel}}$ & 39.0 & 40.4   & \textbf{40.5}   & 40.4   & 40.2   \\
    \hline
    \end{tabular}
\end{minipage}
\end{minipage}

\subsubsection{Study of the Threshold $\Theta$ for POS Algorithm.} We conduct an experiment to validate the optimal value of $\Theta$ for POS, with results shown in \cref{tab6:thred}. As partial regions may have various OAR with ground truths, POS may ignore some partial regions if $\Theta$ is set too high, and may erroneously suppress true positives if $\Theta$ is set too low. Therefore,  we use $\Theta=0.5$ for our experiments.

\begin{table}
\centering
\caption{Effectiveness of the threshold $\Theta$ for partial object suppression (POS) algorithm.}
\label{tab6:thred}
\begin{tabular}{c|ccccccc}
\hline
$\Theta$ & \makebox[0.08\textwidth][c]{1.0}& \makebox[0.08\textwidth][c]{0.9} &\makebox[0.08\textwidth][c]{0.8} & \makebox[0.08\textwidth][c]{0.7} & \makebox[0.08\textwidth][c]{0.6} & \makebox[0.08\textwidth][c]{0.5} & \makebox[0.08\textwidth][c]{0.4} \\
\hline
 $\text{mAP}_{50}^{\text{novel}}$ & 39.4   & 40.2   & 40.2   & 40.4   & \textbf{40.5}   & \textbf{40.5}   & 40.1   \\
\hline
\end{tabular}
\end{table}
\vspace{-15pt}

\subsection{Visualisation Results}

\begin{figure}
  \centering
  \includegraphics[width=\textwidth]{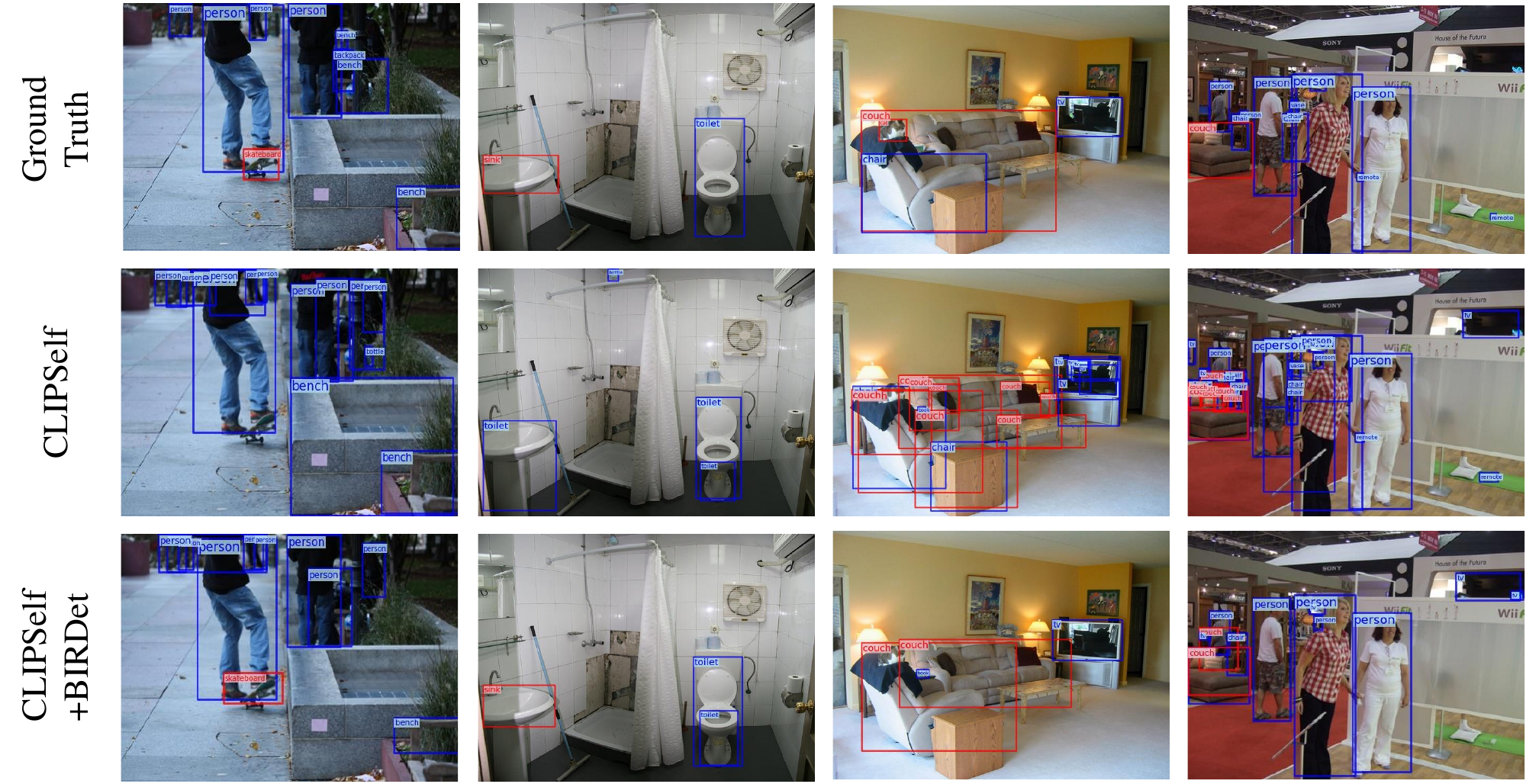}
  \caption{Examples of qualitative results on OV-COCO. Blue detection results are base categories, while red detection results are novel categories. Our analysis demonstrates that BIRDet effectively reduces the misclassification of novel objects as either background or base categories and produces fewer redundant detections compared to the base detector (\ie CLIPSelf~\cite{wu2023clipself}).}
  \label{fig:5}
\end{figure}

In this section, we present the visualization results of BIRDet based on ViT-B CLIPSelf model~\cite{wu2023clipself} and compare it with CLIPSelf. As shown in \cref{fig:5}, our model outperforms CLIPSelf by generating fewer background samples and achieving more accurate detection of novel objects, thus it results in a reduced false positive rate and enhanced detection accuracy.

\section{Conclusion and Limitations}

Currently, open-vocabulary detectors based on CLIP face challenges in handling background samples, particularly for novel categories that are inaccessible during training. In this paper, we introduce BIRDet, an approach designed to  reduce false positive detections caused by background region samples, especially for oversized regions and partial regions. For oversized regions, BIRDet uses scene information to construct background embeddings and compute similarities between scene information and object categories to address bias introduced by scene information. In addition, BIRDet employs the POS algorithm to remove partial regions. Our experiments on OV-COCO and OV-LVIS demonstrate that our approach enhances various open-vocabulary detectors. However, we primarily explore the relationship between background and foreground, leading to a relatively limited improvement when the misclassification rates among categories are relatively high. We hope to overcome these limitations in the future work.

\subsubsection{\ackname}
This work was supported in part by the NSFC under Grant 62206288.

%
%
%
%

\bibliographystyle{splncs04}
\bibliography{7444}
\end{document}